\documentclass[conference]{IEEEtran}

\usepackage{soul}
\usepackage{circledsteps}
\usepackage{cite}
\usepackage{amsmath,amssymb,amsfonts}
\usepackage{graphicx}
\usepackage{xcolor}
\usepackage{textcomp}
\usepackage{booktabs}
\usepackage{multirow}
\usepackage{url}
\usepackage[bookmarks=false]{hyperref}
\usepackage{enumitem}
\usepackage{tikz}
\usepackage[normalem]{ulem}
\usepackage{arydshln}
\usepackage[T1]{fontenc}
\usepackage{microtype}
\usetikzlibrary{shapes.geometric, arrows.meta, positioning, fit,
                  backgrounds, calc, decorations.pathreplacing}

\graphicspath{{figures/}}

\begin{document}
\bstctlcite{IEEEexample:BSTcontrol}

\title{TelcoAgent: A Scalable 5G Multi-KPM Forecasting With 3GPP-Grounded Explainability}

\author{
\IEEEauthorblockN{Geon Kim$^{\dagger}$, Dara Ron$^{\ast}$, Sukhdeep Singh$^{\ddagger}$, Suyog Moogi$^{\ddagger}$, Pranshav Gajjar$^{\ast}$, \\ V V N K Someswara Rao Koduri$^{\ddagger}$, Een Kee Hong$^{\dagger}$ and Vijay K. Shah$^{\ast}$}
\IEEEauthorblockA{$^{\ast}$\textit{NextG Wireless Lab}, North Carolina State University, USA}
\IEEEauthorblockA{$^{\dagger}$Kyung Hee University, South Korea}
\IEEEauthorblockA{$^{\ddagger}$Samsung R\&D Institute India, Bangalore, India}
\thanks{This work was carried out at NextG Wireless Lab, NC State University where Geon Kim was a visiting PhD researcher in Spring 2026.}
}

\maketitle

\begin{abstract}
 Key Performance Measurement (KPM) forecasting is essential for proactive network management of 5G and next-generation telecom networks. However, existing machine learning (ML) approaches face significant limitations in scalability and explainability, restricting their effectiveness in real-world deployments. We propose TelcoAgent, a foundation model-based framework that enables accurate, scalable, and explainable forecasting of multiple KPMs across diverse network cells without the need for site-specific training. Specifically, the framework comprises three key components: (i) an automated three-agent pipeline that constructs a 3rd Generation Partnership Project (3GPP) knowledge graph directly from specification documents, (ii) a scalable, time-series foundation model (TSFM)-based prediction pipeline to deliver accurate, zero-shot forecasting, 
 and finally (iii) a reasoning and explanation pipeline that provides actionable, domain-grounded diagnostics. Evaluated using a $3$-month, city-scale 5G KPM dataset which is based on a U.S.-based network operator, TelcoAgent demonstrates high forecasting accuracy for all $7$ considered  KPMs per cell across $200$ cells, while delivering explainable insights and actionable instructions to address network degradations.

\end{abstract}


\section{Introduction}
\label{sec:intro}

The Federal Communications Commission (FCC) Working Group on Artificial Intelligence (AI), Machine Learning (ML), Testing, and Softwarization highlights forecasting of key performance measurements (KPMs) as a critical capability for network monitoring, proactive orchestration, and control \cite{fcc_report}. Accurate KPM forecasting enables networks to dynamically adapt to service demands and optimize resource utilization, such as avoiding significant energy waste by powering down idle base stations.


Prior literature has applied a range of classical ML algorithms to network KPM forecasting. For instance, regression-based neural networks have been utilized to forecast UE downlink throughput and data traffic~\cite{mostafa2022downlink}. To capture broader dependencies, spatiotemporal graph neural networks are frequently employed for modeling inter-cell KPM interactions~\cite{lin2024multi}. In addition, recurrent architectures are heavily utilized to track temporal changes, including standard recurrent neural networks~\cite{karami2022deeprankpi} and dedicated LSTM models configured for predicting downlink PRB utilization~\cite{rahman2024leveraging}.


Despite recent advances, classical machine learning models face fundamental bottlenecks in accuracy, scalability, and explainability. They struggle to capture nonlinear cross-KPM dependencies, while their reliance on cell-specific training creates severe computational overhead that hinders network-wide scalability. Crucially, these models lack domain-grounded reasoning and merely output predictions without diagnosing root causes, leaving operators without the actionable insights needed for proactive management.

To overcome these limitations, foundation models offer a new paradigm for network management. Time-series foundation models (TSFMs) provide zero-shot capabilities to forecast multiple KPMs across hundreds of cells without retraining, while large language models (LLMs) act as reasoning agents that leverage 3GPP knowledge graphs to transform forecast data into insights.



We introduce TelcoAgent, a reasoning-based LLM framework for large-scale, diverse KPM forecasting and proactive orchestration. It integrates three core modules: (1) an automated 3GPP-based knowledge graph construction pipeline, (2) a TSFM-based zero-shot prediction pipeline, and (3) a \underline{Re}asoning and \underline{Act}ing (ReAct) based explanation pipeline for cross-channel insighs. Our main contributions are as follows:

\begin{itemize}[leftmargin=*,nosep]


    \item We introduce a novel zero-shot multi-KPM forecasting framework that unifies a cross-channel TSFM with 3GPP knowledge retrieval using a ReAct-based agent. This approach enables joint forecasting across diverse KPMs without task-specific training.
    
    \item To enable domain-grounded reasoning, we construct a 3GPP knowledge graph using an automated three-agent pipeline that extracts, aligns, and evaluates knowledge from 3GPP specifications. By linking KPMs to their definitions, formulas, and causal factors, this structured knowledge base provides a robust foundation for precise network analysis.
    
    \item We evaluate TelcoAgent using a 5G dataset which is based on U.S. operator\footnote{The data is syntehsized based on the real-world dataset which mimics the trends of the dataset}. Experimental results show that the prediction pipeline achieves high forecasting accuracy, while the explanation pipeline uses sensitivity analysis to identify root causes across channels, delivering evidence-backed recommendations for network management.
\end{itemize}

\section{Related Work}
\label{sec:related}

\textbf{KPM Prediction in 5G Networks:} Reliable KPM forecasting is a cornerstone of proactive, zero-touch network management. LSTM-based architectures have been widely deployed to forecast traffic and throughput across heterogeneous network slices~\cite{tran2023mlkpi}, while GNN-based approaches exploit the spatial dependencies inherent in base station topologies~\cite{yaqoob2022gnn}. Despite these advances, such supervised paradigms exhibit fundamental limitations, as they demand extensive labeled data, require frequent retraining under distribution shifts, and often overlook cross-KPM correlations.

\textbf{Time-Series Foundation Models:} TSFMs are large pre-trained models capable of forecasting unseen time series without task-specific training. For instance, Chronos-2~\cite{ansari2025chronos2} and Moirai~\cite{woo2024moirai} forecast multiple channels simultaneously in a single inference, whereas MOMENT~\cite{goswami2024moment} processes each channel independently. Although these models achieve strong zero-shot accuracy on general benchmarks, they lack two capabilities essential for network operations. Specifically, they fail to incorporate the domain-specific causal relationships defined in 3GPP standards, and they cannot provide structured rationales for their predictions.

\textbf{LLM Agents and Knowledge-Grounded Reasoning:} LLMs like GPT-4~\cite{achiam2023gpt} enable agent frameworks such as ReAct~\cite{yao2023react}, which interleave reasoning traces with tool-calling to verify information. In telecommunications, benchmarks like ORAN-Bench-13K~\cite{gajjar2025oranbench} provide grounding for 3GPP and O-RAN standards, while OG-RAG~\cite{sharma2025ograg} shows ontology-grounded retrieval enhances factual accuracy over standard RAG. However, these paradigms focus on textual retrieval, lacking mechanisms to bridge domain knowledge with time-series dynamics. TelcoAgent addresses this gap by coupling a TSFM backbone with a 3GPP-grounded knowledge graph, enabling knowledge-grounded forecasting alongside causal insights and actionable recommendations.


\section{TelcoAgent: LLM-based Framework for Telecom Networks}
\label{sec:architecture}

TelcoAgent is a reasoning-based LLM framework designed to deliver explainable KPM forecasting for complex telecom networks. By coupling predictive modeling with domain-grounded reasoning, it provides operators with actionable, instruction-driven guidance. This synergy ensures proactive network management and sustained quality of service even in complex network environments.    

\begin{figure}[t]
  \centering
  \includegraphics[width = 0.48 \textwidth]{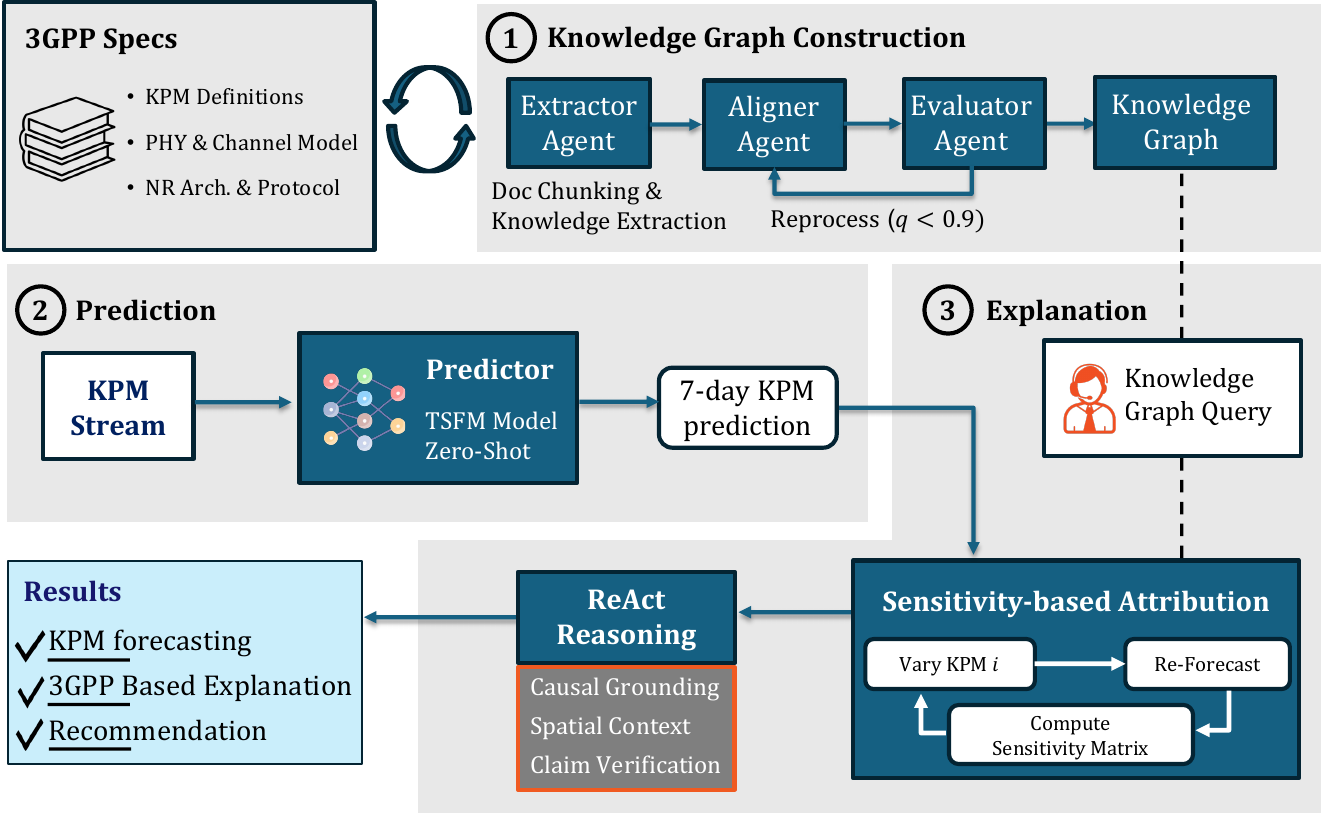}
  \caption{The TelcoAgent architecture comprises three pipelines: {\small \Circled{1}} knowledge graph construction, {\small \Circled{2}} TSFM-based zero-shot forecasting, and {\small \Circled{3}} reasoning and explanation for causal KPM insights and actionable recommendations across multiple cells.}
  \label{fig:architecture}
\end{figure}

TelcoAgent operates through three integrated pipelines as illustrated in Fig.~\ref{fig:architecture}. First, {\small \Circled{1}} \textbf{knowledge graph construction pipeline} extracts structured insights from 3GPP specifications. Second, the {\small \Circled{2}} \textbf{prediction pipeline} leverages TSFMs for zero-shot forecasting across multiple KPMs. Finally, the {\small \Circled{3}} \textbf{explanation pipeline} integrates these forecasts with 3GPP-grounded reasoning to deliver actionable insights. By adopting the ReAct paradigm, TelcoAgent ensures zero-shot adaptability and evidence-backed insights.



\subsection{Knowledge Graph Construction}
\label{sec:graphrag}

\begin{figure}[t]
  \centering
  \fbox{\includegraphics[width=0.9 \columnwidth, trim=7 5 25 23, clip]{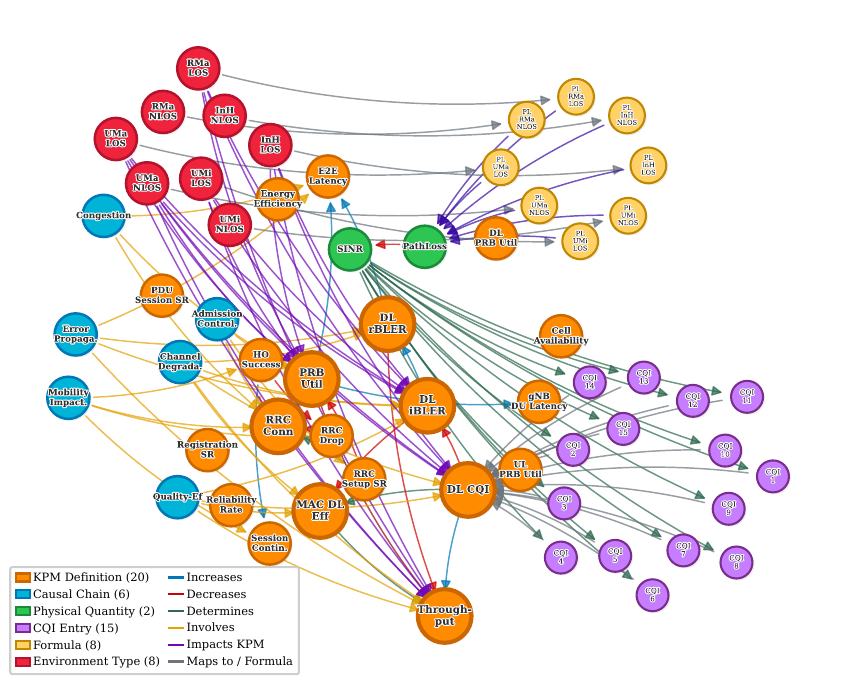}}
  \caption{The constructed 3GPP knowledge graph derived from specifications, encoding KPM definitions, causal chains, and physical-layer constraints for structured network reasoning.}
  \label{fig:knowledge_graph}
  \vspace{- 0.2 in}
\end{figure}

TelcoAgent grounds its reasoning in a 3GPP knowledge graph derived from thirteen specifications to provide a comprehensive domain-specific understanding. TS 28.552, 28.554, and 38.314 define performance counters and measurement-to-KPM derivation chains, while physical layer details come from TS 38.211 to 38.215 and TR 38.901. Architectural and protocol standards such as TS 38.300, 38.321, 38.322, and 38.331 establish structural network dependencies. This knowledge maps into an ontology-based schema illustrated in Fig.~\ref{fig:knowledge_graph}, encompassing KPM definitions, physical-layer constraints, and causal relationships.


Three specialized LLM agents construct the knowledge graph through a sequential pipeline. First, the extractor agent parses 3GPP specifications into section-level chunks to extract {subject, predicate, object} triples, capturing KPM relationships and causal chains. Next, the aligner agent maps these entities to the canonical 3GPP ontology while normalizing inconsistent terminology into single nodes. Finally, the evaluator agent assigns a confidence score $q \in [0, 1]$, triggering a feedback loop for re-alignment if $q$ falls below a threshold $q_{TH}$ to ensure structural and semantic consistency.

\subsection{Prediction Pipeline}
\label{sec:inference}

The prediction pipeline relies entirely on a TSFM to generate robust multi-step KPM forecasts without any domain-specific fine-tuning. A sliding window continuously aggregates historical observations over an interval $L$ to form an input matrix $\mathbf{X} \in \mathbb{R}^{L \times C}$, where $C$ represents the KPM channels. Through zero-shot inference, the TSFM jointly processes this matrix to produce a forecast $\hat{\mathbf{Y}} \in \mathbb{R}^{H \times C}$ over a horizon $H$, naturally capturing inter-KPM dependencies to comprehensively project future performance. By harnessing the generalized representation power of the TSFM, the framework captures complex degradation patterns, establishing a quantitative foundation before proceeding to the explanation pipeline.

\subsection{Explanation Pipeline}
\label{sec:re}

The forecasted values and confidence intervals from the TSFM are passed to a ReAct-based reasoning agent as a structured prompt for causal analysis. The pipeline begins by extracting key metrics—including means, trend slopes, baselines, and percentage changes—directly from the forecasted trajectories. By quantifying the magnitude and velocity of these shifts, the agent establishes a precise quantitative baseline for the cell’s future state, ensuring the diagnostics are grounded in the specific deployment scenario.



To uncover the underlying dynamics driving these forecasted trends, the pipeline systematically quantifies inter-KPM dependencies. Because TSFMs operate as highly non-linear black boxes, the pipeline employs PAX-TS~\cite{kreuzer2025paxts}, a model-agnostic approach that systematically varies source KPMs to compute a cross-channel sensitivity matrix $\mathbf{S} \in \mathbb{R}^{C \times C}$. To resolve the lack of causal directionality in empirical sensitivity scores, the pipeline retrieves directed causal paths from the 3GPP knowledge graph, explicitly matching them with standardized protocol loops.


Building on these directed dependencies, the pipeline isolates the primary drivers of potential network anomalies. This is achieved by synthesizing the variation-based sensitivity scores, the retrieved 3GPP causal chains, and spatial context to capture geographical influences. This multi-modal reasoning allows the pipeline to jointly identify the anomaly trigger, explain the responsible RAN function, and assess the environmental impact, while systematically filtering out spurious, non-causal correlations.

Once the root cause is isolated, the pipeline formulates actionable recommendations to preemptively address the forecasted degradation. Instead of generic monitoring advice, it targets specific RAN parameters by grounding each proposed action in a traceable evidence chain. This targeted guidance enables immediate and precise network configuration adjustments, ensuring that every recommendation is directly mapped back to the underlying sensitivity scores and initial forecast deviations to guarantee technical accuracy.


Finally, the pipeline incorporates an automated self-verification step to ensure absolute reliability and mitigate LLM hallucinations. Before outputting the final insights, a verification module cross-checks all extracted numerical values, including sensitivity scores and KPM statistical descriptors, against the reference computations from the prediction and variation stages. This strict validation guarantees that the generated insights and recommendations rest entirely on verified pipeline-derived data rather than hallucinated text.


\section{Experimental Evaluation}
\label{sec:evaluation}


In collaboration with a U.S.-based network operator, we collected real-world KPMs from a 5G cellular network. This dataset spans diverse geographical regions and traffic scenarios, ensuring a highly representative sample of operational conditions. It is used to validate city-scale KPM forecasting and comprehensively evaluate the evidence-based explanation and recommendation capabilities of our framework. By leveraging such authentic empirical data, we demonstrated the applicability of our framework in handling network anomalies.


\subsection{Experimental Setup}
\label{sec:eval_setup}

\subsubsection{5G KPM Dataset} 
We collected real-world KPMs from a 5G network operating in the Personal Communications Service band of 1850 to 1990 MHz. The dataset was synthesized from KPMs collected continuously from 200 cells of a real network via the centralized Element Management System (EMS) over three months, from September to November 2025, at a fixed one-hour granularity.


\subsubsection{Multi-KPM Forecasting Models} 
The proposed TelcoAgent architecture enables network operators to seamlessly integrate advanced forecasting models into the TSFM prediction module without structural modifications. By decoupling the reasoning engine from predictive algorithms, the framework ensures long-term applicability and protects the core diagnostic logic from disruptions during necessary upgrades. Furthermore, such flexibility is essential for managing the highly dynamic traffic variations inherent to modern cellular deployments. To benchmark this architecture, we evaluate six candidate models comprising three TSFMs and three supervised models.


For supervised baselines, we evaluate three standard models trained per-station on an $81$-day historical dataset. These models serve as independent benchmarks to compare against our zero-shot TSFM module and are not part of the TelcoAgent architecture. First, N-BEATS~\cite{oreshkin2020nbeats} is used to break down KPM patterns into trend and seasonality components. Second, the Gated Recurrent Unit (GRU)~\cite{cho2014gru} is employed to track long-term temporal changes in network traffic while maintaining computational efficiency. Finally, a simple Multi-Layer Perceptron (MLP) acts as a basic reference by directly mapping past observations to future forecasts.

\begin{figure}[!t]
  \centering
  \includegraphics[width = 0.9\columnwidth]{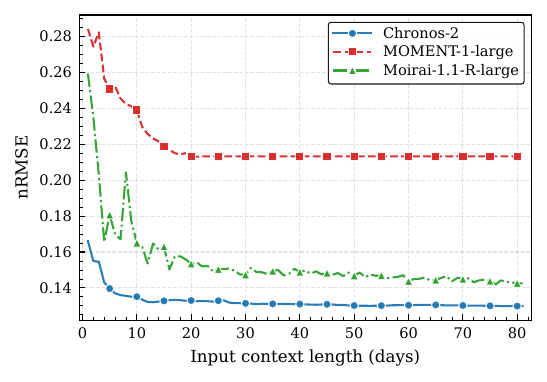}
  \caption{Effect of input context length on average forecasting nRMSE for PRB utilization across 200 cells for three representative TSFMs.}
  \label{fig:input}
\end{figure}

Within the zero-shot framework, we evaluated three representative TSFMs to forecast a $7$-day horizon without site-specific fine-tuning. To maximize predictive accuracy, we conducted a comprehensive sweep over various context windows, selecting the optimal input length for each model based on the nRMSE trends in Fig.~\ref{fig:input}: 81 days for Chronos-2~\cite{ansari2025chronos2}, 81 days for Moirai~\cite{woo2024moirai}, and 22 days for MOMENT~\cite{goswami2024moment}. Both Chronos-2 and Moirai handle all $7$ KPM channels jointly to capture inter-metric correlations, utilizing their respective universal forecasting and any-variate designs. In contrast, MOMENT serves as a baseline by processing each KPM independently. This data-driven selection ensures each model operates under its most favorable configuration, providing a robust analysis of zero-shot scalability for telecom traffic.

\subsubsection{3GPP Knowledge Graph Construction Threshold} 
The threshold $q_{TH}$ is chosen to balance graph density and connection reliability. If $q_{TH}$ is excessively high, the number of edges drops sharply, creating a sparse graph that misses key 3GPP causal links and limits the agent's reasoning capacity. Conversely, a lower threshold permits low-quality connections, introducing noise and computational overhead as the graph becomes unnecessarily dense. Therefore, setting $q_{TH} = 0.9$ establishes an optimal balance; it ensures high relationship quality while preserving enough connectivity to generate reliable, evidence-based recommendations. Ultimately, this filtering prevents the agent from being overwhelmed by irrelevant noise, guaranteeing that the final insights are both accurate and actionable.


\subsubsection{Metrics}
We assess forecasting performance using the normalized Root Mean Squared Error (nRMSE) and Mean Absolute Scaled Error (MASE), averaged over 200 cells. While nRMSE evaluates error magnitude relative to the observed range, MASE measures accuracy relative to a seasonal naive baseline, following the formulations in \eqref{eq:nrmse} and \eqref{eq:mase}:
\begin{subequations}
\begin{equation}
    \text{nRMSE} = \frac{\sqrt{\frac{1}{H}\sum_{t=T+1}^{T+H} (y_t - \hat{y}_t)^2}}{y_{max} - y_{min}}
    \label{eq:nrmse}
\end{equation}
\begin{equation}
    \text{MASE} = \frac{\frac{1}{H}\sum_{t=T+1}^{T+H} |y_t - \hat{y}_t|}{\frac{1}{T-m}\sum_{t=m+1}^{T} |y_t - y_{t-m}|}
    \label{eq:mase}
\end{equation}
\end{subequations}
where $y_t$ and $\hat{y}_t$ are the observed and predicted values at time $t$, $H$ is the forecasting horizon, $T$ is the observation window length, and $m$ is the seasonal period, which we set to 24 to capture the daily cycle.

\begin{table}[!t]
\centering
\caption{Forecasting performance averaged over 200 cells.
\textbf{Bold}: best; \underline{underline}: second best.}
\label{tab:main_results}
\renewcommand{\arraystretch}{1.05}
\resizebox{\columnwidth}{!}{%
\setlength{\tabcolsep}{3pt}
\begin{tabular}{ll rrrrrrr}
\hline\hline
\textbf{Method} & \textbf{Metric} &
\textbf{RRC} & \textbf{CQI} & \textbf{iBLER} & \textbf{rBLER} &
\textbf{MAC Th} & \textbf{PRB} & \textbf{IP Th} \\
\hline
\multirow{2}{*}{Chronos-2} & nRMSE & \textbf{0.12} & \textbf{0.19} & \textbf{0.16} & \textbf{0.14} & \textbf{0.16} & \textbf{0.13} & \textbf{0.15} \\
 & MASE & \textbf{0.76} & \textbf{0.81} & \textbf{0.71} & \textbf{0.67} & \textbf{0.74} & \textbf{0.72} & \textbf{0.72} \\
\hdashline
\multirow{2}{*}{Moirai-1.1-R-base} & nRMSE & \underline{0.13} & \underline{0.20} & \underline{0.16} & \underline{0.14} & \underline{0.16} & \underline{0.14} & \underline{0.15} \\
 & MASE & \underline{0.86} & \underline{0.85} & \underline{0.74} & \underline{0.70} & \underline{0.78} & \underline{0.79} & \underline{0.76} \\
\hdashline
\multirow{2}{*}{MOMENT-1-large} & nRMSE & 0.24 & 0.21 & 0.18 & 0.16 & 0.18 & 0.21 & 0.18 \\
 & MASE & 1.99 & 0.95 & 0.87 & 0.85 & 0.88 & 1.46 & 1.02 \\
\hdashline
\multirow{2}{*}{N-BEATS} & nRMSE & 0.16 & 0.23 & 0.20 & 0.20 & 0.19 & 0.16 & 0.17 \\
 & MASE & 0.96 & 0.94 & 1.00 & 1.26 & 0.91 & 0.94 & 0.88 \\
\hdashline
\multirow{2}{*}{MLP} & nRMSE & 0.19 & 0.26 & 0.22 & 0.23 & 0.21 & 0.19 & 0.19 \\
 & MASE & 1.16 & 1.09 & 1.17 & 1.49 & 1.04 & 1.15 & 1.02 \\
\hdashline
\multirow{2}{*}{GRU} & nRMSE & 0.18 & 0.26 & 0.23 & 0.20 & 0.22 & 0.20 & 0.20 \\
 & MASE & 1.06 & 1.09 & 1.13 & 1.25 & 1.07 & 1.10 & 1.03 \\
\hline\hline
\end{tabular}%
}
\vspace{-0.2in}
\end{table}


We evaluate explanation quality using two complementary metrics, Faithfulness and Answer Relevancy, bounded within [0, 1]. Faithfulness quantifies the proportion of generated 
claims that are factually correct under 3GPP standards, assessing factual reliability and mitigating LLM hallucinations. Answer Relevancy measures the alignment between the generated explanation and the forecasting query on four anomaly-specific axes, ensuring the insights are practically useful for addressing specific network anomalies. These metrics are formally expressed in \eqref{eq:faithfulness} and \eqref{eq:relevancy}:

{\small
\begin{subequations}
\begin{align}
    \text{Faithfulness} &= \frac{1}{|\mathcal{C}(R)|} 
        \sum_{c \in \mathcal{C}(R)} s_{\text{fc}}(c) \label{eq:faithfulness} \\[1.5ex] 
    \text{Answer Relevancy} &= \frac{1}{|E|} \sum_{e \in E} 
        \frac{1}{4} \left( a_{\text{kpm}}^e + a_{\text{time}}^e + 
        a_{\text{mech}}^e + a_{\text{mit}}^e \right) \label{eq:relevancy}
\end{align}
\end{subequations}
}

where $R$ is the generated explanation report, $\mathcal{C}(R)$ is the set of atomic claims extracted from $R$, and $E$ is the set of anomaly events. $\tilde{s}_{\text{f}}(c) \in [0, 1]$ is the normalized faithfulness score for a claim $c$, and $\tilde{a}^e_{\text{kpm}}, \tilde{a}^e_{\text{time}}, \tilde{a}^e_{\text{mech}}$, and $\tilde{a}^e_{\text{mit}} \in [0, 1]$ are the four normalized alignment axes for an event $e \in E$, representing KPM coverage, time window, causal mechanism, and operator action, respectively. All scores are issued by an LLM judge as an integer from 1 to 5, which is then mapped to one of the discrete values $\{0, 0.25, 0.5, 0.75, 1\}$.

\subsection{City-Scale KPM Forecasting}
\label{sec:eval_accuracy}

\begin{figure}[!t]
\centering
\includegraphics[width=0.8\columnwidth]{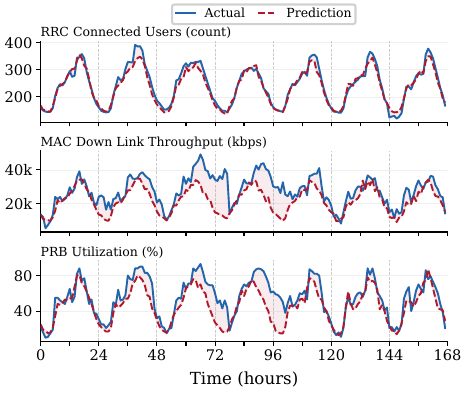} 
\vspace{-0.2 in}
\caption{Comparison of ground truth versus forecasted KPMs by TelcoAgent (employing Chronos-2 as the prediction pipeline) on an hourly basis over 7 days across 200 cells.}
\label{fig:dataset_plots}
\end{figure}

Table~\ref{tab:main_results} reports the per-KPM forecasting performance averaged over 200 cells. Chronos-2 achieves the lowest nRMSE and MASE across all seven KPMs, outperforming both zero-shot and supervised baselines. Fig.~\ref{fig:dataset_plots} illustrates how Chronos-2 accurately forecasts and tracks KPMs on an hourly basis over 7 days, enabling network operators to monitor traffic patterns and proactively orchestrate network resources. This consistent gain across diverse traffic profiles demonstrates the robust generalization of our zero-shot approach.

The performance disparity among zero-shot models offers further insight into their underlying architectures. While Chronos-2 and Moirai consistently rank first and second across all KPMs, MOMENT lags notably on RRC and PRB utilization, despite remaining competitive on link-quality metrics such as Channel Quality Indicator (CQI) and Block Error Rate (BLER). Given that MOMENT processes each channel independently, this performance gap suggests that capturing cross-channel dependencies is vital for KPMs driven by traffic volume and system load. For these metrics, multivariate correlations between user activity and resource consumption appear to be a critical factor for forecasting accuracy.



\label{sec:eval_ex}

\subsection{Explanation Quality and Causal Grounding}
\label{sec:eval_explanation}

The explanation pipeline produces structured reports
({\small \Circled{A}--\small \Circled{E}}) mapping the operator's decision process from situational awareness to intervention. To verify that these insights remain faithful to established telecom domain knowledge rather than hallucinated content, we evaluated $200$ cells
using ORANSight~\cite{oransight2025} as a domain-specific LLM judge. The results show a mean Faithfulness of $0.615$ and a mean Answer Relevancy of $0.807$ across all evaluated cells, confirming that the reports consistently address complex explanation tasks while remaining anchored in 3GPP-grounded evidence.

\begin{figure}[t]
  \centering
  \includegraphics[width=0.8\columnwidth]{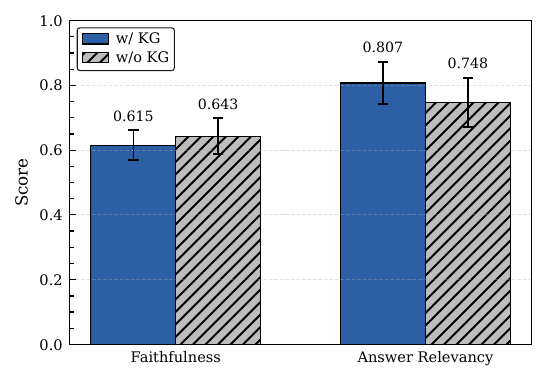}
  \caption{Performance comparison of explanation quality metrics with and without the 3GPP knowledge graph.}
  \label{fig:kg}
\end{figure}

\begin{figure}[t]
  \centering
  \includegraphics[width=\columnwidth]{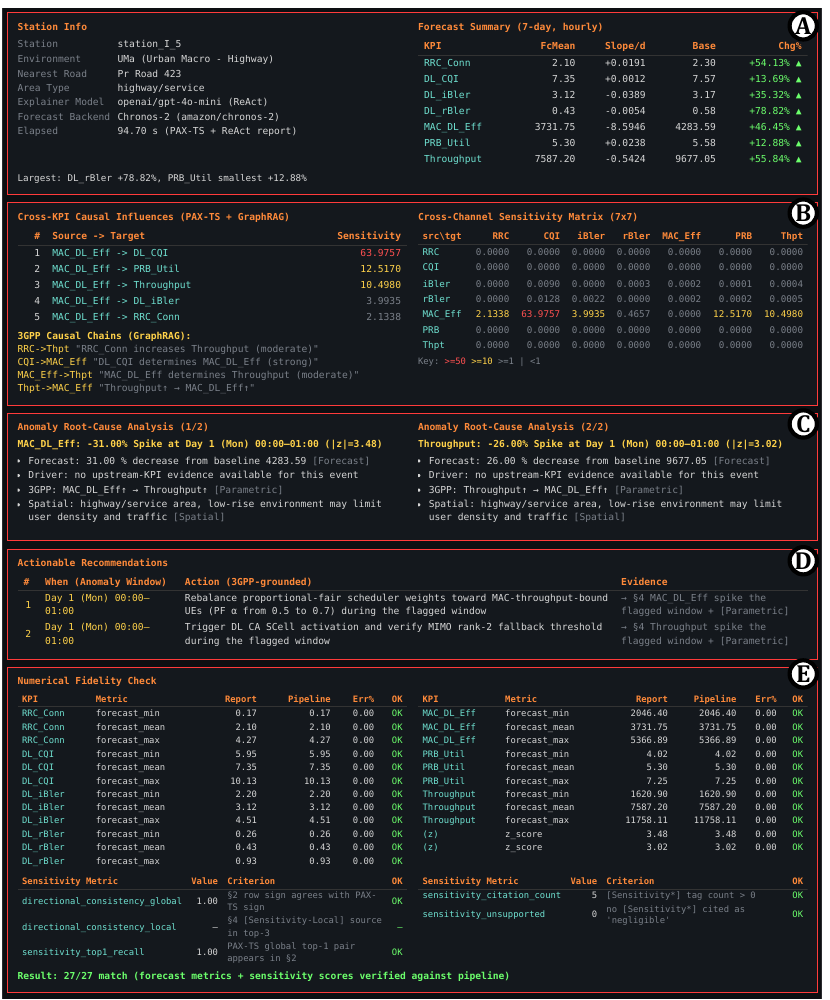}
  \caption{A representative output of the explanation pipeline, structured into five components ({\small \Circled{A}--\Circled{E}}), providing 3GPP-grounded insights and actionable recommendations for a given cell.}
  \label{fig:report}
  \vspace{- 0.2 in}
\end{figure}

As illustrated in Fig.~\ref{fig:kg}, the 3GPP knowledge graph contributes to operational relevance. Removing it drops Answer Relevancy from $0.807$ to $0.748$, representing a $7.4\,\%$ decrease, which indicates that graph-derived causal chains assist in identifying anomaly mechanisms and operator actions. Conversely, utilizing the knowledge graph slightly reduces Faithfulness from $0.643$ to $0.615$ due to minor paraphrase drift when translating rich protocol details into text. Because this variance remains well within one standard deviation, the knowledge graph effectively anchors the agent's reasoning in validated logic to deliver actionable insights while maintaining overall factual accuracy. Fig.~\ref{fig:report} presents a representative output from the explanation pipeline\footnote{Results and the 3GPP knowledge graph of the proposed TelcoAgent are available at \url{https://github.com/NextG-Wireless-Lab-NC-State/TelcoAgent}.}.

{\small \Circled{A}}~summarizes the expected behavior of each KPM over the next seven days based on the deployment scenario, such as Rural Macro (RMa) or Urban Macro (UMa). For every KPM, the report includes the predicted mean, daily trend slope, baseline mean, and percentage change, with directional arrows marking the sharpest baseline deviations. This enables operators to rapidly assess both the magnitude and velocity of performance shifts, facilitating the prioritization of cells likely to experience throughput degradation or resource congestion.



In 5G NR, KPMs are coupled through directed loops, such as link adaptation and scheduling, which symmetric statistics fail to capture. {\small \Circled{B}} quantifies the strongest of these dependencies for a given cell, displaying the top-5 KPM pairs ranked by variation sensitivity alongside a full $7 \times 7$ dependency matrix. For each identified pair, the pipeline retrieves the matching 3GPP causal chain from the knowledge graph. Unlike correlation-based methods, this variation-based approach captures directed causality, enabling operators to trace anomalies back to the specific RAN functions that trigger them.


To explain KPM deviations in {\small \Circled{C}}, the pipeline integrates variation sensitivity scores to identify the primary driver, 3GPP causal chains to reveal protocol mechanisms, and spatial context to capture geographical influences. This combination jointly explains the anomaly trigger, the responsible RAN function, and the environmental impact. The pipeline further ensures reliability by filtering spurious correlations, treating co-varying KPMs with low variation sensitivity as merely coincidental and non-causal events.


Rather than generic advice, each recommendation in {\small \Circled{D}} identifies specific parameters and ranges, such as Outer Loop Link Adaptation (OLLA) targets, maximum Hybrid Automatic Repeat Request (HARQ) limits, and Radio Resource Control (RRC) inactivity timers. By anchoring these suggestions to the predicted anomaly window, the framework enables operators to preemptively schedule adjustments before the onset. This specificity grounds each action in the exact RAN function responsible for the forecasted deviation. Furthermore, an integrated evidence column links recommendations to driving sensitivity scores and forecast changes, ensuring full traceability back to the underlying protocol behavior.

To mitigate LLM hallucinations, {\small \Circled{E}} cross-verifies all numerical values in the report against the actual pipeline outputs. An automated process compares $33$ extracted metrics, including $5$ variation sensitivity scores and $28$ KPM descriptors such as mean, slope, baseline, and percentage change, against reference computations. Across all $200$ cells, $99.8\%$ of these values match within a predefined tolerance. This high level of consistency confirms the precision of the generated reports, ensuring that the recommendations in {\small \Circled{D}} rest on verified numerical evidence.


\section{Conclusion}
\label{sec:conclusion}

We introduced TelcoAgent, an LLM-based framework for joint multi-KPM forecasting and verifiable causal reasoning. By coupling a zero-shot TSFM with an autonomously constructed 3GPP knowledge graph, the framework generates traceable insights that link each forecast to its RAN function. Evaluations across $200$ base stations demonstrated that TelcoAgent outperformed supervised baselines across all $7$ KPMs and achieved $99.8\%$ numerical fidelity. These results indicate that zero-shot scalability and specification-grounded explainability can be integrated without site-specific retraining or manual verification. To enhance reasoning capabilities, future work will expand the knowledge graph by integrating broader specifications. Furthermore, we plan to incorporate spatio-temporal context from neighboring cells to deploy TelcoAgent as an O-RAN rApp for closed-loop resource orchestration.


\section*{Acknowledgment}
This work was supported by the Institute of Information \& Communications Technology Planning \& Evaluation(IITP)-Digital Innovation Talent Short-Term Intensive program grant funded by the Korea government(MSIT)(IITP-2024-RS-2024-00434743) 

\bibliographystyle{IEEEtran}
\bibliography{references/refs}

@IEEEtranBSTCTL{IEEEexample:BSTcontrol,
  CTLuse_forced_etal       = "yes",
  CTLmax_names_forced_etal = "6",
  CTLnames_show_etal       = "1"
}

@techreport{fcc_report,
  author       = {{FCC TAC AI Working Group}},
  title        = {The Importance of Artificial Intelligence and Data for the Telecommunications Industry},
  institution  = {Federal Communications Commission, Technological Advisory Council},
  year         = {2020},
  note         = {Available: \url{https://www.fcc.gov/sites/default/files/fcc_aiwg_2020_whitepaper_final.pdf}},
}

@inproceedings{karami2022deeprankpi,
  author    = {M. Karami and M. R. Tanhatalab and E. Pourmami},
  title     = {{DeepRanKPI}: Time Series {KPI}s Prediction in a Live Cellular Network with {RNN}},
  booktitle = {Proc. IEEE Int. Conf. Green Energy, Comput. Sustain. Technol. (GECOST)},
  year      = {2022},
  pages     = {41--45},
}

@article{lin2024multi,
  author  = {J. Lin and T. Lan and B. Zhang and K. Lin and D. Miao and H. He and J. Ye and C. Zhang and Y.-F. Li},
  title   = {Multi-Scenario Cellular {KPI} Prediction Based on Spatiotemporal Graph Neural Network},
  journal = {IEEE Trans. Autom. Sci. Eng.},
  volume  = {22},
  pages   = {5131--5142},
  year    = {2024},
}

@inproceedings{mostafa2022downlink,
  author    = {A. Mostafa and M. A. Elattar and T. Ismail},
  title     = {Downlink Throughput Prediction in {LTE} Cellular Networks Using Time Series Forecasting},
  booktitle = {Proc. IEEE Int. Conf. Broadband Commun. Next Gener. Netw. Multimedia Appl. (CoBCom)},
  year      = {2022},
  pages     = {1--4},
}

@inproceedings{rahman2024leveraging,
  author    = {M. H. Rahman and A. Jain and A. P. {Da Silva} and S. Singh and M. R. Chowdhury and H. S. Dhillon and M. Hong},
  title     = {Leveraging {O-RAN SC AI/ML} Framework and Non-{RT RIC} for {AI}-Driven Network Slice {QoS} Optimization},
  booktitle = {Proc. IEEE GLOBECOM Workshops},
  year      = {2024},
  pages     = {1--6},
}

@article{achiam2023gpt,
  author  = {{OpenAI}},
  title   = {{GPT-4} Technical Report},
  journal = {arXiv preprint arXiv:2303.08774},
  year    = {2023},
  note    = {[Online]. Available: \url{https://arxiv.org/abs/2303.08774}},
}

@inproceedings{tran2023mlkpi,
  author    = {P. N. Tran and O. Delgado and B. Jaumard and F. Bishay},
  title     = {{ML KPI} Prediction in {5G} and {B5G} Networks},
  booktitle = {Proc. Joint Eur. Conf. Netw. Commun. 6G Summit (EuCNC/6G Summit)},
  year      = {2023},
  pages     = {502--507},
}

@inproceedings{yaqoob2022gnn,
  author    = {M. Yaqoob and R. Trestian and H. X. Nguyen},
  title     = {Data-Driven Network Performance Prediction for {B5G} Networks: A Graph Neural Network Approach},
  booktitle = {Proc. IEEE Int. Conf. Consum. Electron. (ICCE)},
  year      = {2022},
  pages     = {377--382},
}

@article{ansari2025chronos2,
  author  = {A. F. Ansari and O. Shchur and J. K{\"u}ken and A. Auer and B. Han and
             P. Mercado and S. S. Rangapuram and H. Shen and L. Stella and X. Zhang and
             M. Goswami and S. Kapoor and D. C. Maddix and P. Guerron and T. Hu and
             J. Yin and N. Erickson and P. M. Desai and H. Wang and H. Rangwala and
             G. Karypis and Y. Wang and M. Bohlke-Schneider},
  title   = {{Chronos-2}: From Univariate to Universal Forecasting},
  journal = {arXiv preprint arXiv:2510.15821},
  year    = {2025},
  note    = {[Online]. Available: \url{https://arxiv.org/abs/2510.15821}},
}

@misc{kreuzer2025paxts,
      title={{PAX-TS}: Model-agnostic multi-granular explanations for time series forecasting via localized perturbations}, 
      author={Tim Kreuzer and Jelena Zdravkovic and Panagiotis Papapetrou},
      year={2025},
      eprint={2508.18982},
      archivePrefix={arXiv},
      primaryClass={cs.LG},
      url={https://arxiv.org/abs/2508.18982}, 
}

@inproceedings{woo2024moirai,
  author    = {G. Woo and C. Liu and A. Kumar and C. Xiong and S. Savarese and D. Sahoo},
  title     = {Unified Training of Universal Time Series Forecasting Transformers},
  booktitle = {Proc. Int. Conf. Mach. Learn. (ICML)},
  year      = {2024},
}

@inproceedings{yao2023react,
  author    = {S. Yao and J. Zhao and D. Yu and N. Du and I. Shafran and K. Narasimhan and Y. Cao},
  title     = {{ReAct}: Synergizing Reasoning and Acting in Language Models},
  booktitle = {Proc. Int. Conf. Learn. Represent. (ICLR)},
  year      = {2023},
}

@inproceedings{sharma2025ograg,
  author    = {Sharma, Kartik and Kumar, Peeyush and Li, Yunqing},
  title     = {{OG}-{RAG}: Ontology-grounded retrieval-augmented generation for large language models},
  booktitle = {Proc. Conf. Empirical Methods Natural Lang. Process. (EMNLP)},
  year      = {2025},
  pages     = {32962--32981},
  doi       = {10.18653/v1/2025.emnlp-main.1674}
}

@INPROCEEDINGS{gajjar2025oranbench,
  author={Gajjar, Pranshav and Shah, Vijay K.},
  booktitle={Proc. IEEE Consumer Commun. Netw. Conf. (CCNC)}, 
  title={{ORAN-Bench-13K}: An Open Source Benchmark for Assessing {LLMs} in Open Radio Access Networks}, 
  year={2025},
  volume={},
  number={},
  pages={1--4},
  doi={10.1109/CCNC54725.2025.10975994}}

@inproceedings{goswami2024moment,
  author    = {M. Goswami and K. Szafer and A. Choudhry and Y. Cai and S. Li and A. Dubrawski},
  title     = {{MOMENT}: A Family of Open Time-Series Foundation Models},
  booktitle = {Proc. Int. Conf. Mach. Learn. (ICML)},
  year      = {2024},
}

@inproceedings{oreshkin2020nbeats,
  author    = {B. N. Oreshkin and D. Carpov and N. Chapados and Y. Bengio},
  title     = {{N-BEATS}: Neural Basis Expansion Analysis for Interpretable Time Series Forecasting},
  booktitle = {Proc. Int. Conf. Learn. Represent. (ICLR)},
  year      = {2020},
}

@inproceedings{cho2014gru,
  author    = {K. Cho and B. van Merri{\"e}nboer and C. Gulcehre and D. Bahdanau and F. Bougares and H. Schwenk and Y. Bengio},
  title     = {Learning Phrase Representations using {RNN} Encoder--Decoder for Statistical Machine Translation},
  booktitle = {Proc. Conf. Empirical Methods Natural Lang. Process. (EMNLP)},
  year      = {2014},
  pages     = {1724--1734},
  doi       = {10.3115/v1/D14-1179}
}

@ARTICLE{oransight2025,
  author={Gajjar, Pranshav and Shah, Vijay K.},
  journal={IEEE Trans. Mach. Learn. Commun. Netw.},
  title={{ORANSight-2.0}: Foundational {LLMs} for {O-RAN}}, 
  year={2025},
  volume={3},
  number={},
  pages={903--920},
  doi={10.1109/TMLCN.2025.3592658}}

\end{document}